\documentclass[]{article}
\usepackage{proceed2e}
\usepackage{graphicx}
\usepackage{amssymb}
\usepackage{amsmath}
\usepackage{epstopdf}
\usepackage{cite}
\usepackage{listings}
\usepackage{hyperref}
\usepackage{url}
\newcommand{\be}{\begin{eqnarray} \begin{aligned}}
\newcommand{\ee}{\end{aligned} \end{eqnarray} }
\newcommand{\benn}{\begin{eqnarray*} \begin{aligned}}
\newcommand{\eenn}{\end{aligned} \end{eqnarray*} }
\newcommand{\setK}{\mathcal{K}}
\newcommand{\setM}{\mathcal{M}}
\newcommand{\set}[1]{\mathcal{#1}}
\usepackage{color}

\newcommand{\mo}{\mathcal{O}}
\newcommand{\vc}[1]{{\bf #1}}

\title{A Sequence of Relaxations Constraining Hidden Variable Models} 

\author{ {\bf Greg Ver Steeg \and Aram Galstyan} \\  
Information Sciences Insitute \\  
University of Southern California\\ 
\{gregv,galstyan\}@isi.edu\\
} 
 
\begin{document} 
 
\maketitle 
 
\begin{abstract} 
Many widely studied graphical models with latent variables lead to nontrivial constraints on the distribution of the observed variables. 
Inspired by the Bell inequalities in quantum mechanics, we refer to any linear inequality whose violation rules out some latent variable model as a ``hidden variable test'' for that model.  
Our main contribution is to introduce a sequence of relaxations which provides progressively tighter hidden variable tests.
We demonstrate applicability to mixtures of sequences of i.i.d. variables, Bell inequalities, and homophily models in social networks. For the last, we demonstrate that our method provides a test that is able to rule out latent homophily as the sole explanation for correlations on a real social network that are known to be due to influence. 
\end{abstract} 
 
 \section{Introduction}
 
 Bayesian graphical models provide an intuitive framework for modeling dependence among variables that often correspond to relationships observed in the real world. When all variables are observed, the correspondence between graphical models and (non--experimental) probability distributions is completely described in terms of conditional independence relations directly implied by the graph\cite{pearl}. Often, some variables cannot be observed, leading to nontrivial constraints on the distributions over observed variables including non--independence equality constraints\cite{tianpearl,geigermeek2} and inequality constraints\cite{kangtian}. 
Understanding these constraints can allow us to rule out the otherwise difficult to test hypothesis that observed correlations are only due to a common dependence on some unknown variable. 
A method for generating constraints for general latent variable models
would be useful in many contexts.
 
 For instance, recent high profile studies have identified counter-intuitive traits (e.g. obesity \cite{obesity}) as being socially contagious. To take such a claim seriously, one would have to rule out the alternate possibility that some hidden factor causes people to become friends and that this same factor encourages obesity (such a mechanism is referred to as ``latent homophily'').   In fact, a recent paper shows that latent homophily and influence are non-parametrically unidentifiable in social networks\cite{cosma}. 
 Is there a limit to the amount of correlation between friends that can be explained by latent homophily? 
 
This question also calls to mind the Bell inequalities in quantum physics.
They place a bound on how strongly correlated two particles can be if each particle's state is described fully by some hidden variable.
Violation of this bound shows us that our assumption is wrong; the two particles can only be fully described by some joint ``entangled'' state\cite{peres}. 
Our method allows us to construct ``Bell inequalities'' for other quantities like correlations in social networks.
If the correlations between friends are explained by latent homophily, they will obey some bound. Violating this bound allows us to rule out latent homophily as the explanation for correlations. These bounds can be considered tests of the proposed hidden variable model.
   
For any model, consider the set of all observable probability distributions consistent with the model. We use sum-of-squares methods and semidefinite programs to create a sequence of convex relaxations for this set\cite{parrilo}. 
Our ability to create this sequence of relaxations requires only that the observable probabilities can be written as polynomials in terms of the  unknown conditional distributions.
%\footnote{The model parameters themselves must be specifiable as a semi--algebraic set, a simple requirement which we describe later.}
Clearly, this includes all graphical models and many generalizations. 
We focus on progressively tighter approximations of this set because any general, exact method would also be able to generate all Bell inequalities, a problem known to be computationally difficult, discussed in Sec.~\ref{bell}.
The technique is most powerful when the space of observable probability distributions is already convex, so that our sequence of convex relaxations can converge exactly, in principle.
For instance, this is the case for any graphical model which contains a single latent root node connected to any number of observed nodes.
Furthermore, because the latent variable serves only to produce convex combinations of other observable probability distributions, we can ignore it altogether, even though its domain size may be infinite.

We begin with a simple example to develop geometric intuition of the technique. Next we describe tools from algebraic geometry that allow us to produce constraints for a large class of models including latent variable graphical models. 
We then demonstrate some applications focusing on the increasingly significant problem of identifying influence in social networks. 
Finally, we will contrast our technique to previous work in this area.
 
% 
%Like all Bayesian graphical models, they are essentially polynomial relationships between the observed probabilities and the unobserved conditional probabilities, $P(node|parents)$.  What these models have in common is that there is a hidden node, observed probabilities are convex combinations of fully observed dist.
%$$P(abc) = \sum_r P(r) P(abc|r)$$

%Is r a confounder?

%A pedagogical introduction \cite{nielsen} and a more nuanced exploration in \cite{peres}.
% 
% Pearl \cite{pearl} on nontrivial constraints.
% 
% Introduce models?
% 
% Relaxed constraints on these models.
% 
% -more general - semi-algebraic model specification + convexity
\vspace{0.3cm}
\section{Example: Mixture of i.i.d. sequences}\label{example}

Imagine $A = (A_1,\ldots,A_k)$ is a sequence of $k$ binary variables, $A_j \in \{T,H\}$.
The null hypothesis is that each sequence is produced by randomly drawing a weighted coin from a jar, flipping it $k$ times (i.i.d.), and then replacing it. 
Formally, these independence assumptions mean the probability distribution, $P$, can be decomposed as,\footnote{A,B,X,Y,R,E will be used to represent random variables throughout. Instantiations of variables are typically omitted for readability but will be made explicit when necessary. We also use vector notation so that, e.g., $\vc{y}$ is a vector whose components are labeled with $y_i$. }
\be\label{eq:coinflip}
P(A) &=&  \sum_{R=1}^\infty P(R) \prod_{j=1}^k P(A_j | R),
 \ee
 with the additional constraint that 
 for each weighted coin, labeled by $R$, the distribution for the coins is i.i.d., that is,
 $$\forall i,j,j', ~~P(A_j = H | R=i ) = P(A_{j'} = H | R=i ).$$
 Moving to a more algebraic description, we say that a weighted coin comes up heads with probability $P(A_j = H | R=i) \equiv \eta_i \in [0,1]$, and there are a possibly infinite number of weights for $i=1,\ldots,\infty$ each of which occurs with unknown probability $P(R=i) \equiv \lambda_i$.  If we call $h(A)$ the number of heads in a sequence then we can rewrite our distribution as,
 \be
 P(A) &=& \sum_i \lambda_i \eta_i^{h(A)}  (1-\eta_i)^{k-h(A)} .
\ee
We must remember to constrain $\lambda_i,\eta_i \in [0,1]$, and $\sum_i \lambda_i = 1$. What constraints does this model put on the observed distribution, $P(A)$? 

\begin{figure}[t]
\vspace{0.6cm}
\center
\includegraphics[width=6cm]{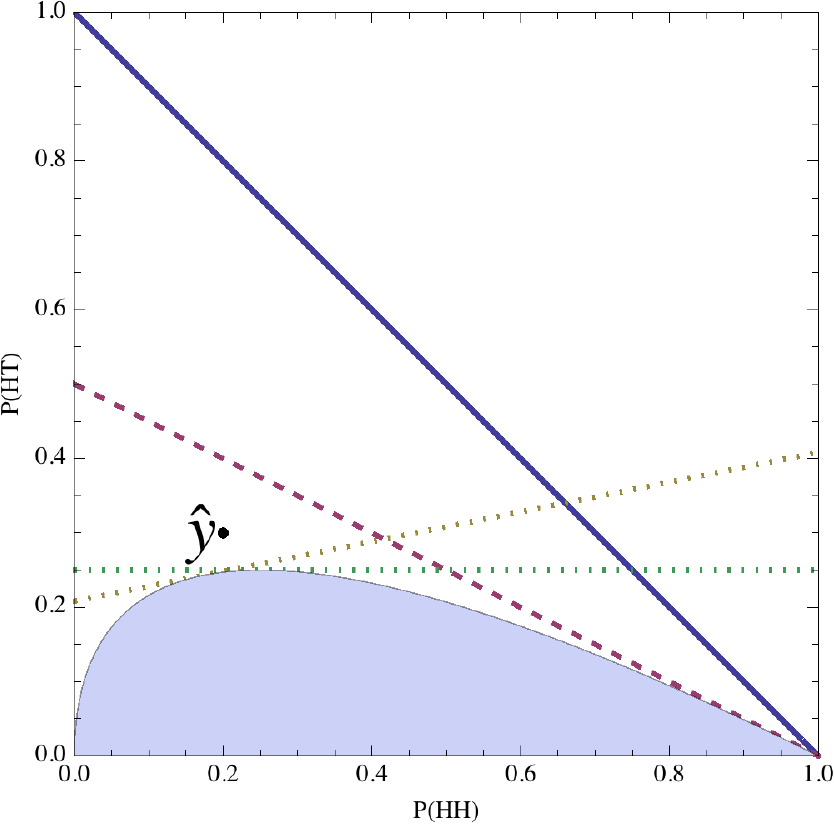}
\caption{The shaded region represents accessible probability distributions according to our model constraints. For contrast, all probabilities in the simplex are below the solid line and all exchangeable distributions are below the dashed line.  For the observed distribution, $\hat{\vc{y}} $, our model is ruled out by either hidden variable test represented by the dotted lines.}
\label{fig:coinflip}
\end{figure}

For a given weighted coin, the distribution of outcomes for flips $i$ and $j$ should be i.i.d., but the unobserved weight serves as a confounder. 
The observation that allowed distributions must be exchangeable, $P(A_1,\ldots,A_k) = P(A_{\pi(1)},\ldots,A_{\pi(k)})$ for any permutation $\pi$, simplifies the problem but is not a sufficient constraint. Methods like implicitization\cite{kangtian2} and quantifier elimination\cite{geigermeek} are prohibited by allowing the size of the latent variable's domain to be infinite. 
%\note{ We might use the trivial converse of de Finetti's theorem to note that allowed distributions must be exchangeable, $P(A_1,\ldots,A_k) = P(A_{\pi(1)},\ldots,A_{\pi(k)})$ for any permutation $\pi$, but this restriction is not sufficient (for finite $k$).}  

Instead, we take a geometric approach, where we start by constructing the vector $\vc{y}$ whose entries are the probabilities $P(A)$ for each $A \in \{H,T\}^k$, and therefore resides in a $2^k$ dimensional simplex. In that case, we note that Eq.~\ref{eq:coinflip} has the form of the convex hull of a surface where each extreme point is specified by the probability distribution for an i.i.d. sequence for a single weighted coin. 

If we take the simple case of two coin flips, $k=2, A\in\{HH,HT,TH,TT\}$, then, to simplify further, we note that, e.g. $P(TH),P(TT)$ can be fixed by the requirements of normalization and exchangeability. This leaves us with
%\be
%y = \left( \begin{array}{c} P(HH) \\ P(TH)\\P(HT)\\P(TT) \end{array} \right) 
%=  \sum_i \lambda_i \left( \begin{array}{c} \eta_i^2 \\ \eta_i (1-\eta_i) \\ \eta_i (1-\eta_i) \\ (1-\eta_i)^2 \end{array} \right).
%\ee
\be
\left( \begin{array}{c} y_1 = P(HH) \\ y_2 = P(HT) \end{array} \right) 
=  \sum_i \lambda_i \left( \begin{array}{c} \eta_i^2 \\ \eta_i (1-\eta_i) \end{array} \right).
\ee
 The shaded region in Fig.~\ref{fig:coinflip} shows the possible probabilities. We refer to extreme points as vectors of the form $(\eta_i^2, (1-\eta_i) \eta_i)$, while other points are formed as convex combinations of these extreme points.
 In this case, we can easily enumerate the remaining constraints on $P(A)$. We note that all extreme points of the distribution have the form $P(HT) = \sqrt{P(HH)} - P(HH)$ (the curved boundary of the shaded region) so we can see that convexity allows all distributions satisfying the conditions $P(HT) \leq \sqrt{P(HH)} - P(HH)$ and positivity. 

We seek a more general approach that will work for high-dimensional examples, even when an exact description of the convex hull is not forthcoming. We note that a convex hull can also be given as an intersection of half--spaces of the form $\{\vc{y}: \vc{b} \cdot \vc{y} \leq c \}$\cite{convexanalysis}. 
Clearly, our simple example is not a polytope, so a finite number of half--spaces will not suffice to describe it.
However, in practice, we do not need an exact description of the convex hull. 
Instead, after observing many sequences of $A_1,\ldots,A_k$, we can determine the distribution of observed variables $\hat{P}(A)$ and we want to test whether this distribution could have been produced by our null model. In our example, this is equivalent to determining if $\hat{P}(A)$ produces a point in the shaded region of Fig.~\ref{fig:coinflip}. If we find our point outside the shaded region, we can rule out our null model as a possible explanation. If, on the other hand, our distribution is inside the shaded region, we can only conclude that our null model is one possible explanation. 

To construct a test capable of ruling out a latent model, we can test membership in any half--space that contains our convex set. In this example that means we are looking for vector $b$, constant $c$, so that 
\benn b_1 \eta^2 + b_2 \eta (1-\eta) \leq c, ~~\forall \eta \in [0,1]. \eenn
If we were to observe $\hat{P}(HH) = \hat{P}(TT) = 0.2,\hat{P}(HT) = \hat{P}(TH) = 0.3$, then $\hat{\vc{y}}  = (0.2,0.3)$, shown in Fig.~\ref{fig:coinflip}. 
Clearly, this point is outside our convex shaded region. 

To find a test demonstrating that $\hat{\vc{y}} $ is outside of our set we want a $\vc{b}$ so that $\vc{b}\cdot \hat{\vc{y}} > c$.
For instance, one can easily verify that $b_1=0,b_2=1,c=1/4$ or $b_1=-1/5,b_2=1,c=5/24$ are such hyperplanes, shown as dotted lines in Fig.~\ref{fig:coinflip}. 
These tests suffice to rule out the null hypothesis that our sequences were produced from an unknown distribution of weighted coins. Of course, if $\hat{\vc{y}} $ is an experimental distribution, it is only an estimate of the true distribution, and we can only rule out the null hypothesis with some confidence.
In this example, we are able to visualize the solution and find simple tests by hand, but generally, the dimensionality of our problems will be large, so we must develop tools to automatically find good tests.

\section{Algebraic geometry}

We consider models where the model parameters may be restricted to any semi--algebraic set, that is, a subset of $\mathbb{R}^m$ defined by a finite number of polynomial equalities and inequalities. For graphical models the ``parameters'' will be conditional probabilities; this use of the word should not be confused with ``parametric models'', which is not what we are considering.
For simplicity, we will consider inequalities only.  We assume, as is the case for graphical models, that equality constraints may be used to simply reduce the overall number of parameters, and constraints will generally include positivity and normalization.
%A semi--algebraic set on $\mathbb{R}^n$ is any set defined by a finite number of polynomial equalities and inequalities. The Tarski-Seidenberg theorem tells us that the projection of a semi--algebraic set is still a semi--algebraic set. 
$$\setK = \{\vc{x} \in \mathbb{R}^m: g_i(\vc{x}) \geq 0, i=1,\ldots, l \}$$

Furthermore, the observable quantities, $\vc{y} \in \mathbb{R}^n$, must be written as a convex combination of polynomials over the model parameters, $f_i(\vc{x}^{j}), \vc{x}^{j} \in \setK$. 
\be\label{algmodel}
y_i = \sum_j \lambda_j f_i (\vc{x}^{j})
\ee
with $\lambda_i \geq 0, \sum_i \lambda_i =1$. 
For graphical models, we consider ``observables'' to be measurable properties of a system like expectation values, e.g., $y_1 = \langle \delta_{A_1,H} \rangle = P(A_1=H)$. Whereas, the $\lambda_i$ will represent probabilities of latent variables which we are marginalizing out.

We call the set of possible $\vc{y}$, given the model constraints, $\setM$. Notice that Eq.~\ref{algmodel} can be represented as the convex hull of a set ($\mbox{conv}$), that is, all possible convex combinations of a set of extreme points,\footnote{Restrictions on the $\lambda_i$ in Eq.~\ref{algmodel} may lead to a smaller set of possible observables $\setM' \subseteq \setM$.  In what follows, we will construct a sequence of outer relaxations that converge towards $\setM$, so if $\setM$ is already a relaxation of $\setM'$, we will be ultimately limited in how well we can approximate it.}
\be
\setM =  \mbox{conv}(\{\vc{y} \in \mathbb{R}^n : \exists \vc{x} \in \setK, \vc{y} = \vc{f}(\vc{x}) \})
\ee
An alternate representation of this convex set is given in terms of the intersection of half--spaces containing it\cite{convexanalysis}.
\be\label{SLHset}
\set{B} = \{\vc{b} \in \mathbb{R}^n: \forall \vc{x} \in \setK, \vc{y}=\vc{f}(\vc{x}), 1 - \vc{b} \cdot \vc{y} \geq 0 \}\\
\setM =  \mbox{int}(\set{B}) \equiv \{\vc{y} \in \mathbb{R}^n : \forall \vc{b} \in \set{B}, \vc{b} \cdot \vc{y} \leq 1\}.
\ee
Note that this formulation implicitly presupposes that the origin is inside our convex set. This condition can be insured with a simple translation of the vector $\vc{y}$. E.g. $\vc{y} \rightarrow \vc{y} - \int_K d\vc{x} \vc{f}(\vc{x})/ \int_K d\vc{x}$. 
Because the half--space described by each $\vc{b} \in \set{B}$ contains all the extreme points of the set, $\vc{y}=\vc{f}(\vc{x}), \forall \vc{x} \in \setK $, it also contains the convex hull of these points. 

Given a representation of the convex hull, $\set{B}$, one can determine that a point $\hat{\vc{y}} $ is outside $\setM$ by finding a $\vc{b}\in \set{B}$ such that $\vc{b} \cdot \hat{\vc{y}} > 1$. That is, by showing that $\hat{\vc{y}} $ is not in the intersection of half--spaces comprising $\setM$. 
If we consider a subset $\set{RB} \subset \set{B}$, this set amounts to a convex relaxation on the original set $\setM$. That is,$$\vc{b} \in RB \wedge \vc{b} \cdot \hat{\vc{y}}  > 1 \rightarrow \hat{\vc{y}}  \notin \setM,$$ but for this relaxation, the converse is not true. 

{\bf SOS Relaxations}
To construct a subset of $\set{B}$ which can be efficiently described and optimized over,
we will need some standard results about positive and sum-of-squares (SOS) polynomials. See \cite{parrilo} for a review of the large body of work about SOS and positive polynomials and their relationship to semi-definite programming. For completeness, we summarize the key ideas here.

Eq.~\ref{SLHset} describes $\set{B}$ as all $\vc{b}$ so that $1-\vc{b}\cdot \vc{f}(\vc{x}) \geq 0$, for $\vc{x} \in \setK$. In other words, for some polynomials in $\vc{x}$, parametrized by $\vc{b}$, we want only the polynomials that are non-negative on $\setK$. When dealing with positive polynomials, the simplest relaxation is to instead consider 
bounded degree sums-of-squares polynomials\cite{parrilo}
\benn
SOS_d = \{s(\vc{x}): \exists q_i(\vc{x}) \in \mathbb{R}[\vc{x}], \mbox{deg}(q_i(\vc{x})) \leq d/2, \\ 
s(\vc{x}) = \sum_i q_i(\vc{x})^2\}.
\eenn
By construction, polynomials in $SOS_d$ are guaranteed to be non-negative.
If we can write $1-\vc{b}\cdot \vc{f}(\vc{x}) = s_0(\vc{x}) $, for $s_0 \in SOS_d$, we can guarantee $1-\vc{b}\cdot \vc{f}(\vc{x}) \geq 0$. 
\footnote{In general, it is not true that every positive polynomial can be written as an SOS.}

Efficient computational methods using SOS polynomials are enabled by the fact that they can be written in the form $$s(\vc{x}) = \vc{z}^\intercal A \vc{z},$$ where $\vc{z}=(1,x_1,x_1 x_2,x_1 x_2^2 ...)$ is a vector of monomials in the variables and $A\succeq 0$ indicates a positive semidefinite matrix. Then our condition for $1-\vc{b}\cdot\vc{f}(\vc{x}) = s_0(\vc{x})$ amounts to linear relationships between coefficients along with a linear matrix inequality, $A\succeq 0$. These types of problems are called semidefinite programs(SDP) and many powerful techniques exist to solve them. 

In our case, because we only demand positivity on a bounded region $\setK$, defined by polynomials $g_i(\vc{x}) \geq 0$, we make things a little easier. The set of all polynomials positive on $\set{K}$ is called the ``positive cone'' of $\setK$ and includes SOS polynomials by default. 
Roughly, this is just the set of polynomials that are formed as sums of products of the $g_i(\vc{x})$ and $s(\vc{x}) \in SOS$. Clearly, if $g_i(\vc{x}) \geq 0, \forall \vc{x}\in \setK \rightarrow s(\vc{x}) g_i(\vc{x}) \geq 0,\forall \vc{x}\in \setK $ and $s(\vc{x}) g_1(\vc{x}) g_2(\vc{x}) \geq 0,\forall \vc{x}\in \setK $, and so on. 

Therefore, we define the set, $\set{RB}_1 \subseteq  \ldots \subseteq \set{RB}_d \subseteq \set{B}$.
\benn
\set{RB}_d = &\{\vc{b} \in \mathbb{R}^n : \forall \vc{x} \in K, s_i(\vc{x}) \in SOS_d, \\
&1 - \vc{b} \cdot \vc{f}(\vc{x}) = s_0(\vc{x}) + \sum_i s_i(\vc{x}) g_i(\vc{x}) \}
\eenn
By construction, $s_0(\vc{x}) + \sum_i s_i(\vc{x}) g_i(\vc{x}) \geq 0$ for all $x\in \setK$. For any $\vc{b}\in \set{RB}_d$, this proves $1 - \vc{b} \cdot \vc{f}(\vc{x}) \geq 0$, and for any $\vc{y}$ in the convex hull of $\vc{f}(\vc{x})$, this will also be true. 
This amounts to a sequence of convex relaxations of the set $\setM$. 
$$\setM = \mbox{int}(\set{B}) \subseteq \mbox{int}(\set{RB}_d) \ldots \subseteq \mbox{int}(\set{RB}_1)$$
Exactly as defined, we have not included all polynomials in the positive cone of $\setK$(we excluded terms with products of $g_i(\vc{x})$ for computational ease), so this sequence does not necessarily converge to $\setM$. Including these terms can provide theoretical, if computationally impractical, guarantees of convergence\cite{convergence}. 

For a specific observed distribution $\hat{\vc{y}} $, we search for a hyperplane $\vc{b} \in \set{RB}_d$ so that $\vc{b} \cdot \hat{\vc{y}} $ is maximized.
\be\label{sosprogram}
\max_{\vc{b},s_i(\vc{x})} \vc{b}\cdot \hat{\vc{y}} && \\
1 - \vc{b} \cdot \vc{f}(\vc{x}) -\sum_i s_i(\vc{x}) g_i(\vc{x}) &=& s_0(\vc{x})\\
s_i(\vc{x}) &\in& SOS_d
\ee
This format corresponds to a SOS program and it can be efficiently translated into a semidefinite program and solved numerically\cite{parrilo}. 
We use SOSTools\cite{sostools} in MATLAB to convert SOS programs to SDP which are then solved by, e.g., SeDuMi\cite{sedumi}, example code is available\cite{SM}. In practice, even very large SDPs can be solved efficiently. Unfortunately, it is difficult to construct rigorous bounds on their complexity. 

%%SUPPLEMENTAL MATERIAL
For the example in Sec.~\ref{example}, $\vc{x} \in \mathbb{R}^1$ and corresponds to the variable we called $\eta,$ the weight of the coin. On the other hand, $\hat{\vc{y}} \in \mathbb{R}^2$, because we have two observables in this case. To check if $\hat{\vc{y}}$ were produced according to the model described, we would solve Eq.~\ref{sosprogram} with 
$\hat{\vc{y}} = (0.2,0.3), g_1(x) = x (1-x), f_1(x) = x^2, f_2(x) = x (1-x). $ The value of $d$  can be increased to produce tighter bounds.
Solving this SOS is constructive in that it finds $b$ and specific SOS polynomials proving that $1 - \vc{b} \cdot \vc{y} \geq 0$, for any $y\in \setM$.
Furthermore, if we find that $\vc{b} \cdot \hat{\vc{y}}   > 1$, this constitutes proof that the statistics $\hat{\vc{y}} $ could not have been generated by our model.

\section{Application to graphical models}

A probabilistic graphical model, or Bayesian network, consists of a directed acyclic graph where each node is a variable, and the edges reflect the relationships between these variables. The graph has a rigorous mathematical interpretation which can be described via conditional independence relations among the variables, or, equivalently, by a specific decomposition of the probability distribution in terms of conditional probability distributions. For a review, see \cite{pearl}. Although these graphs can also be given a causal interpretation, allowing the prediction of some variables given that others are fixed (an intervention), we will not consider interventional distributions here.

\subsection{Bell inequalities}\label{bell}

\begin{figure}[ht]
\vspace{0.6cm}
\center
\includegraphics[width=4cm]{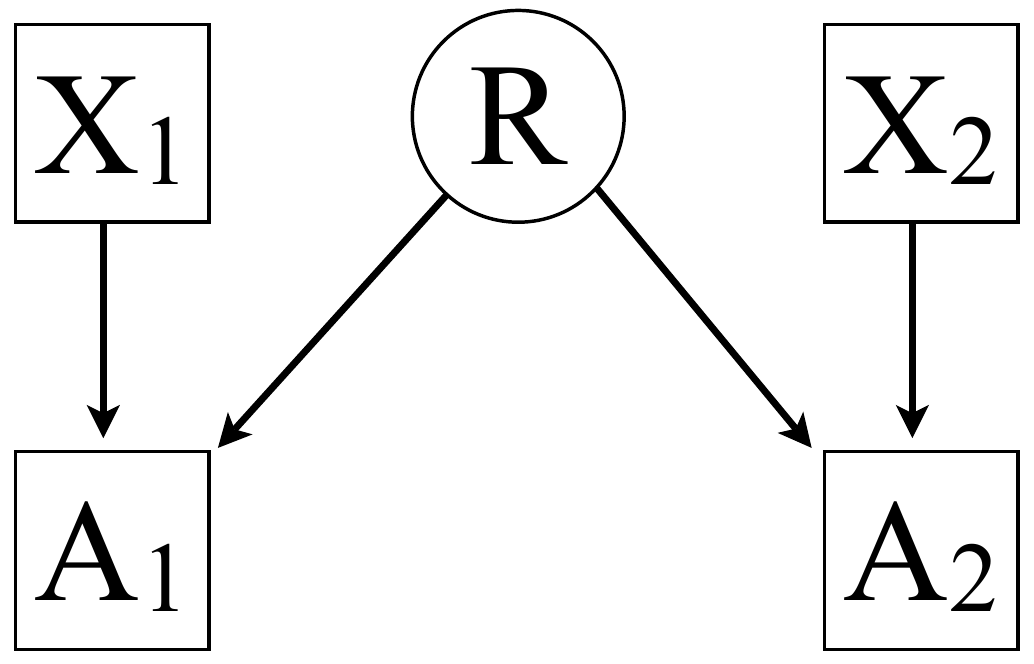}
\caption{Graphical model describing the assumptions of ``local realism'' in a CHSH experiment. Each outcome $A_i$ depends only on the local measurement choice $X_i$ and some common hidden variable $R$.}
\label{fig:chsh}
\end{figure}

Quantum physicists rarely appreciate the fact that Bell inequalities are a special case of inequality constraints for a latent variable graphical model. Bell inequalities refer generally to any tests of ``local realism'' in quantum physics, while we will consider a specific realization known as the CHSH inequality. See \cite{nielsen} for a pedagogical introduction and \cite{peres} for a more nuanced exploration of the meaning of such tests (see App.~\ref{a1} for a primer to the alternate formalism). Briefly, we imagine two detectors, $i=1,2$ that are separated so that no signals can pass between them. Each detector can make only one of two possible measurements $X_i = \{0,1\}$ resulting in a binary outcome, $A_i = \{0,1\}$. The assumptions of local realism are summarized in Fig.~\ref{fig:chsh}. 
Basically, we assume that party $i$'s measurement outcome depends only on her own measurement choice(``local'') and some hidden variable(``realism'').
$$
P(A_1,A_2|X_1,X_2) = \sum_R P(R) P(A_1 | X_1,R) P(A_2 | X_2, R)
$$
This formula has the form required by Eq.~\ref{algmodel}. This example has already been thoroughly studied; its structure is known to be a convex polytope whose nontrivial facets are given by inequalities of the form \cite{nielsen},
\be\label{chsh}
\sum_{A_i,X_i \in \{0,1\}} P(A_1,A_2|X_1,X_2) \delta_{A_1\oplus A_2, X_1 \cdot X_2} \leq 3, 
\ee
where $\oplus$ is defined as addition modulo 2.
We can reproduce this well-known bound in a novel way by solving Eq.~\ref{sosprogram}.

Practically, for any equivalent experimental setup (two observers who have two measurement choices resulting in one of two outcomes), we can measure the quantity on the LHS of Eq.~\ref{chsh}. For any model of the form in Fig.~\ref{fig:chsh}  we expect this quantity to be bounded by 3. In particular, if the two parties are space-like separated (no signal from one party could reach the other in time to have an effect), all models obeying classical, relativistic physics have that form.  If, on the other hand, we measure a value larger than 3, we can rule out this model. 
This is precisely what has been done and is one of the strongest pieces of evidence in favor of quantum theory. 
The maximum value achievable from measuring entangled quantum particles is $\approx 3.4$\cite{nielsen}.

We could extend the model in Fig.~\ref{fig:chsh} to more parties or more measurement settings. The space of observed distributions will still be described by a convex polytope, but it will have many facets and it has been shown that deciding membership in these polytopes is NP complete\cite{pitowsky}. This shows that any general, exact method for deciding whether a distribution is compatible with a latent variable graphical model will be computationally difficult, at least in some cases. This motivates our approach to instead look for a sequence of progressively tighter relaxations. 

%\subsection{Stationary Markov sequences}

\subsection{Latent homophily models}\label{models}

Sociologists often observe that individuals who are connected in a social network exhibit behaviors that are highly correlated. This correlation is usually explained via two effects: homophily and influence. Influence, or contagion, supposes that actors change to become more similar to their neighbors in the network.
Whereas, homophily posits that individuals form connections in the network precisely because they are already similar.
%Mention external causation?

Suppose Alice is friends with Bob, a smoker, and some time later Alice begins smoking. If Alice would not have begun smoking if she had not known Bob, we would certainly say she was influenced by Bob. 
An alternate explanation is that Alice and Bob both enjoy high risk lifestyles, and that is why they became friends. 
Alice is already predisposed to start smoking, and would have begun even if she had never met Bob. 
A typical sociological study would attempt to control for this covariate -- either by measuring Alice and Bob's risk-taking tendencies or some substitute that indicates those tendencies. In this case, the difficulty comes from trying to measure all possibly relevant covariates; this is the approach taken in \cite{obesity,aral}.

\begin{figure}[ht]
\vspace{0.6cm}
\center
\includegraphics[width=4.5cm]{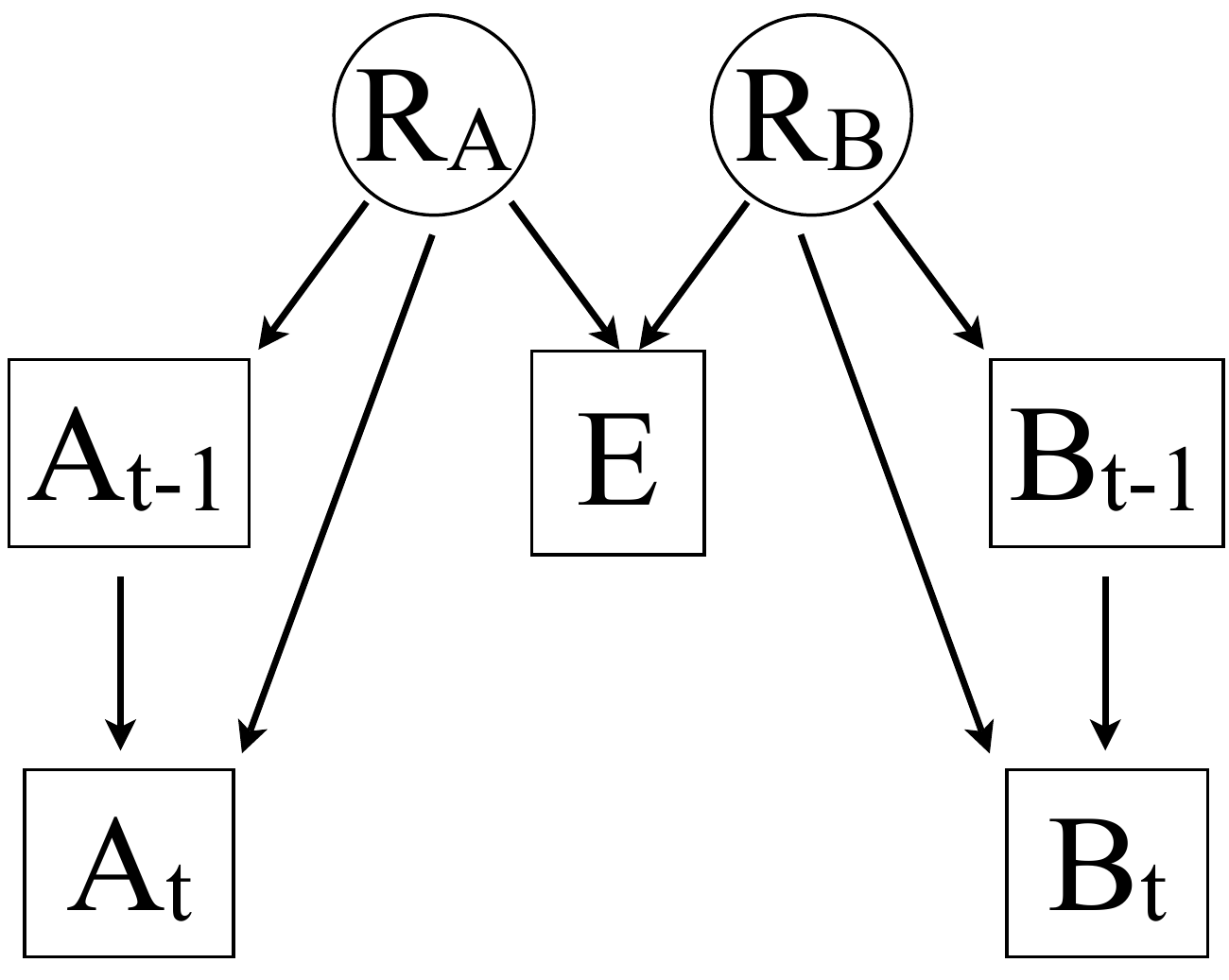}
\caption{A slice of a latent homophily model. We observe a sequence of actions for $A_1,\ldots,A_T$ (sometimes abbreviated $A$) and $B$ that depend on some hidden attributes $R_A,R_B$. Presence and properties of edges between them, $E$, depend in some arbitrary way on $R_A,R_B$.}
\label{homophily}
\end{figure}

 In Fig.~\ref{homophily}, we start with the most general picture of latent homophily. We have two actors Alice(A) and Bob(B) whose actions we observe at various time steps, $t=1,\ldots,T$. We consider some hidden attributes of Alice($R_A$) and Bob($R_B$) and $E$ depends somehow on both hidden attributes and represents
information about edges between them (e.g., a time-dependent sequence of edges, possibly directed or weighted, possibly including edges of various kinds). 
Unlike previous works\cite{anagnostopoulos,neville}, we do not assume that $E$ depends symmetrically on $R_A$ and $R_B$, an important consideration in networks with asymmetric (directed) links. 
%Although we do not explicitly include some observed attributes in this model as in\cite{cosma}, their presence makes no difference to the results. 

%\begin{figure}[ht]
%\vspace{0.6cm}
%\center
%\includegraphics[width=7cm]{homophily.pdf}
%\caption{Model of latent homophily: We observe a sequence of actions for $A$ and $B$ that depend on some hidden attributes $R_A,R_B$. Presence and properties of edges between them, $E$, depend in some arbitrary way on $R_A,R_B$.}
%\label{homophily}
%\end{figure}

Given $E$, what correlations are possible between $A$ and $B$? Below we use the definition of the graphical model\cite{pearl} in Fig.~\ref{homophily} along with some simple manipulations using Bayes' rule.  
%\be\label{eq:jointprob}
%P(A_{1:T},B_{1:T}|E) &=& \sum_{R_A,R_B}  P(A_{1:T}|R_A) P(B_{1:T}|R_B)  P(E|R_A,R_B) P(R_A) P(R_B) /P(E) \\
% &=&  \sum_{R_A,R_B}  P(A_{1:T}|R_A)  P(B_{1:T}|R_B)   P(R_A,R_B | E) 
%\ee
%We also take into account that $A_t$ may depend on $A_{t-1}$ in addition to $R_A$.
\be\label{eq:jointprob}
P(A_{1:T},B_{1:T}|E) =  \sum_{R_A,R_B} P(R_A,R_B | E) \times \\
 \prod_t  P(A_t | A_{t-1}, R_A) P(B_t | B_{t-1}, R_B) 
\ee

As written, the correlations possible between $A$ and $B$ are not very restricted. In fact, \cite{cosma} have shown that it is not possible to distinguish between correlations from latent homophily, given by the structure of Fig.~\ref{homophily}, and influence, represented by a direct link from $B_{t-1}$ to $A_t$ when an edge from $B$ to $A$ exists. Part of this difficulty is due to the fact that the transition probabilities $P(A_t | A_{t-1},R_A)$ may change at each time step, allowing essentially arbitrary dynamics. We show that eliminating this freedom allows us to find tests that rule out latent homophily.
%It is easy to see that an arbitrary marginal distribution $\bar{p}(A_{1:T},B_{1:T})$ can be written in a manner consistent with this graphical model,\footnote{In fact, in a future longer version of this work we will demonstrate that $|\mbox{dom}(R)| \sim 2^T$ is a necessary and sufficient condition to reproduce an arbitrary correlation between $A$ and $B$.} given appropriate definition of $P(R)$. Let $R$ be a vector of $2^{2T}$ values so that each is associated with a pair of sequences for Alice and Bob, $R = (A,B)$ .
% Now take 
%\be\label{arbitrary}
%P(R=(A,B)|E) = \bar{p}(A,B) \\
%P(A_t | A_{t-1},R=(A',B')) = \delta_{A_t,A'^t}
%\ee
%and similarly for B.
%Plugging these values into Eq.~\ref{eq:jointprob}, we reproduce the arbitrary distribution $\bar{p}$.
%
%Intuitively, we have reproduced arbitrary correlations by making the space of our hidden attribute large and then allowing the dependence of $A_t$ on $R$ to change at each time step. Effectively, this is the same as allowing the hidden attribute to fluctuate with time. Generally, when we say that actions depend on hidden attributes, this is not what we mean. We assume that the attributes (e.g. gender, IQ, etc.) are not changing (or at least not quickly) with respect to the observed actions (smoking, posting on a social network, etc.), and we wish to explain the latter in terms of the former. Therefore, in the next section, we restrict ourselves to models with this property.

{\bf Static latent homophily model}
We now define static latent homophily models (SLH) by demanding the crucial addition of stationarity: the transition probability does not change over time.
\be\label{stationary}
\forall t,t',r:~~~ &P(A_{t} = a | A_{t-1} = b, R = r)& \\
= ~&P(A_{t'}=a | A_{t'-1}=b, R=r)&.
\ee
The homophily model of \cite{cosma} also looks like Fig.~\ref{homophily}, but without the stationary assumption in Eq.~\ref{stationary}. 
%Note that the demonstration in Eq.~\ref{arbitrary} that latent homophily reproduces arbitrary correlations only holds without stationarity.

The stationary Markov assumption restricts the probability of observing certain sequences. If Bob's state is a sequence of coin flips, it is highly unlikely that Alice independently produces the same sequence without seeing (or being influenced by) Bob's coin flips. We will make this intuition more precise in the next section.

\subsubsection{Algebraic geometry of SLH}\label{geometry}

Looking at Eq.~\ref{eq:jointprob}, we can see that we have just defined a polynomial mapping from the small space of conditional probabilities to the larger space of Alice and Bob's observed joint probability distribution. The structure of Eq.~\ref{eq:jointprob} is a convex combination over the (possibly infinite) factorizable joint distributions.

For simplicity, we now consider a SLH where we restrict ourselves to variables $A_t,B_t \in \{\pm 1\}$, and we have conditioned on some arbitrary value for $E$ (e.g., $E=1$ for a directional link from A to B).
Each variable sequence, $A_{1:T}$ is a Markov chain with associated transition probabilities that depend on the unknown value of $R_A$. We denote by $\alpha_+ (\alpha_-)$ the probability that $A$ flips from $+ (-)$ to $- (+)$ at some time step and $\alpha_0 = P(A_1 = +1)$. We have similar parameters for $B: \beta_{+,-,0}$.
For legibility below, we suppress the functional dependence of these probabilities on $R_A$ and we take $A$ to represent the sequence $A_{1:T}$. 
\be\label{eq:marginal}
P(A_{1:T} | R_A) 
=& \alpha_+^{F_+(A)} \alpha_-^{F_-(A)} \\
&(1-\alpha_-)^{S_-(A)} (1-\alpha_+)^{S_+(A)} \\
&\alpha_0^{1/2(1+A_1)} (1-\alpha_0)^{1/2(1-A_1)} 
\ee
\benn
F_{\pm} (A) = \sum_{t=1}^{T-1} \frac14 (1 \pm A_t) (1 - A_{t+1} A_t) \\
S_{\pm} (A) = \sum_{t=1}^{T-1} \frac14 (1 \pm A_t) (1 + A_{t+1} A_t)
\eenn
The same equations hold replacing $A$ with $B$ and $\alpha$ with $\beta$. 

We can define the parameter vector, $$\vc{x} = (x_1,\ldots,x_6) \equiv  (\alpha_+,\alpha_-,\alpha_0,\beta_+,\beta_-,\beta_0).$$ 
Now consider the expected outcomes from some arbitrary set of measurements $$y_j \equiv \langle \mo_j(A,B) \rangle, j=1,\ldots,n.$$ In principle, the set of $\mo_j(A,B)$ could consist of the indicator functions for each possible outcome (in which case $n=2^{2T}$), but we would like to reserve the ability to pick a smaller set of measurements for computational reasons later on.
Setting $R=(R_A,R_B)$, then 
%$$P(A,B|E) = \sum_R h_{AB} (x_R) P(R|E),$$
\be\label{polymap}
 f_j(x^R)  &\equiv& \sum_{A,B} P(A_{1:T} | R_A) P(B_{1:T} | R_B) \mo_j(A,B) \\
y_j  &=& \sum_{R} P(R|E) f_j (x^R)
\ee
% where $h_{AB}$ represents a polynomial in $x$ of maximum degree $2 T$, defined by Eqs.~\ref{eq:jointprob},\ref{eq:marginal}. 
This represents a polynomial mapping from $\mathbb{R}^6 \rightarrow \mathbb{R}^n$ where the domain is the region $$\setK = \{\vc{x} \in \mathbb{R}^6: g_i(\vc{x}) = x_i (1-x_i) \geq 0, i=1,\ldots,6 \}$$ because each $x_i$ represents a different transition (or prior) probability. The set of all $\vc{y}$ is just the convex hull of $\vc{f}(\vc{x})$ where $\vc{x}\in \setK$.

%Now we are ready to define the set of interest. 
%\be
%SLH &=& \mbox{conv}(  \{y \in \mathbb{R}^n :  \exists x \in K, \vc{y}=\vc{f}(\vc{x})) \} )
%\ee
% If and only if $y \in SLH$, there exists 

%A dual description of a convex set is as an intersection of half-spaces defined by some (possibly infinite) set of hyperplanes.  We only need to ensure that the extreme points of $\vc{y}$ are in the half-space.
\subsubsection{Results for a synthetic example}

\begin{figure}[ht]
\vspace{0.6cm}
\center
\includegraphics[width=6cm]{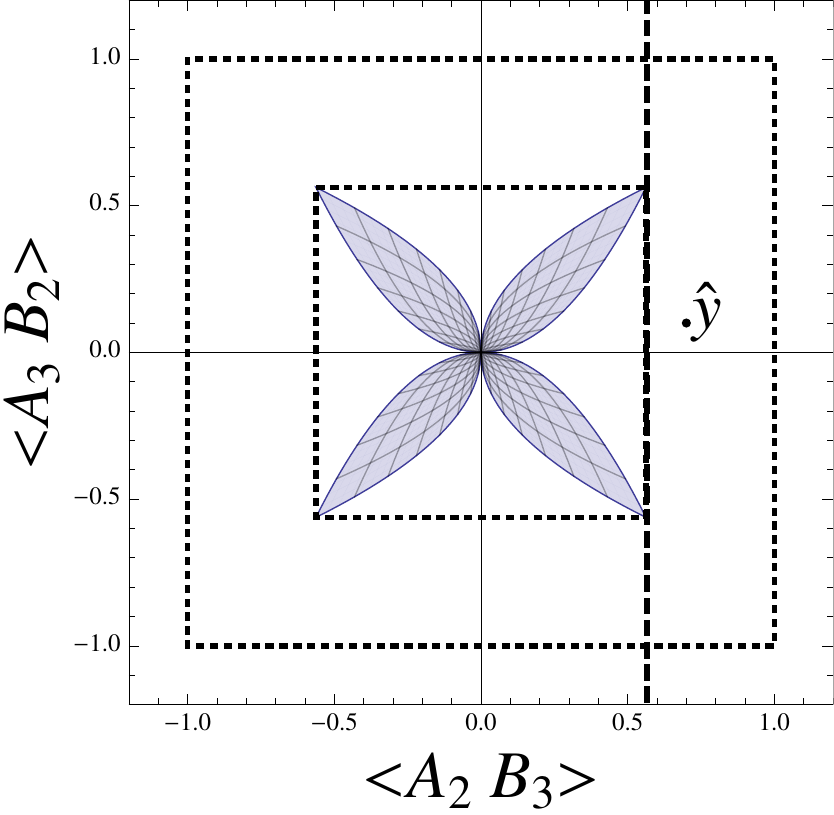}
\caption{ The outer dotted box corresponds to the allowable region for arbitrary probability distributions. The convex hull of the shaded region, denoted by the inner dotted box, corresponds to allowable distributions for this simple model. Solving Eq.\ref{sosprogram} for the point $\hat{\vc{y}} $ returns the dashed line.}
\label{example_bound}
\end{figure}

We begin by considering a simple example which is possible to visualize completely. We observe only two statistics, the correlation between Alice's state at $t=2$ with Bob's state at $t=3$ and vice versa. 
We set$$(y_1,y_2) = (\langle A_2 B_3 \rangle,\langle A_3 B_2\rangle),$$ 
We also constrain the model parameters, by saying the Alice and Bob's states are described by symmetric, stationary Markov chains, and we arbitrarily fix their initial states to be $-1$ with probability $1/8$.
$$\alpha = \alpha_+ = \alpha_-, \beta = \beta_+ = \beta_-, \alpha_0 = \beta_0 = 1/8.$$ 
Even so, for each of the possibly infinite values the latent variable may take, Alice and Bob may have a different transition probability.
\be
y_1 &=&  \sum_R P(R) 9/16 (1 - 2 \alpha(R)) (1 - 2 \beta(R))^2 \\
y_2 &=&  \sum_R P(R) 9/16 (1 - 2 \alpha(R))^2 (1 - 2 \beta(R))
\ee
The space of admissible $\vc{y}$ for this model (with $\alpha,\beta \in [0,1]$) is the convex hull of the shaded region in Fig.~\ref{example_bound}. Whereas, for an arbitrary distribution, any point in the outer dotted box is attainable. We want to test if, e.g. $\hat{\vc{y}}  = (0.7,0.1)$ is in the convex hull of the shaded region, and therefore an allowed distribution for this model, by solving Eq.~\ref{sosprogram}. In this case, we find $\vc{b} = (16/9,0)$, so that $\vc{b}\cdot \hat{\vc{y}}  \approx 1.24 > 1$, along with polynomials $s_0(\alpha,\beta),s_1(\alpha,\beta),s_2(\alpha,\beta) \in SOS_d$ (omitted for brevity) so that 
\benn
1 - 16/9 \cdot 9/16 (1 - 2 \alpha) (1 - 2 \beta)^2 \\= s_0(\alpha,\beta)+s_1(\alpha,\beta) \alpha(1-\alpha)+s_2(\alpha,\beta)\beta (1-\beta) \\
\geq 0~~~~~ \forall \alpha,\beta \in [0,1].
\eenn
In this example, we recover a facet of the convex hull exactly. MATLAB code for this example is provided in \cite{SM}.

\subsubsection{Results on Digg social network}

By construction, our test is only able to rule out latent homophily as the explanation for correlations. 
Assuming that correlations are produced by some alternate model, the test could fail to rule out latent homophily as an explanation for two reasons. First, it could be that the alternate model produces correlations that are actually the same as those produced by latent homophily (unidentifiable). Second, it could be that the alternate model produces correlations that are impossible for a latent homophily model, but our relaxation is so loose that it nevertheless includes these correlations. 
In App.~\ref{a2}, we consider an artificial model of influence and deduce the minimum strength of influence necessary for latent homophily to be ruled out. Below, we demonstrate the usefulness of our test for a real social network.

We started with a real world social network from the online news portal ``digg.com'' \cite{lermandigg}. This network had $M=1,731,659$ edges and $N=279,634$ nodes.
As in \cite{neville}, we do a semi-synthetic analysis by simulating a known influence model on the real graph of this social network. 
For our influence model we started all the nodes in a random state $\pm 1$. At each step, we picked a random pair of nodes $A,B$ who are connected by a (directed) edge from $A \rightarrow B$ and had $B$ copy $A$'s state. Then we considered three time slices from this evolution to construct the statistics $\hat{P}(A_{1:3}, B_{1:3}|E=1)$, where $E=1$ means there exists a directed edge from $A$ to $B$. The time between observations should be chosen long enough so that most nodes have changed, which we accomplished by setting the observations $M$ steps apart.

For all results below, we set $\mo_j(A,B)$ to consist of the indicator functions for each possible outcome. Solving Eq.~\ref{sosprogram}, using $d=3$ and $\hat{P}$ as our estimate of the unknown true distribution $P^*$, returned an observable $\mo(A,B) = \sum_i b_i \mo_i(A,B)$, so that $\forall P \in SLH, \langle \mo \rangle_{P} \leq 1$, while $\bar{\mo} = \langle \mo \rangle_{\hat{P}} = 1.15$. Additionally, the range of $\mo$ is needed, so we found that $\mo \in [-17,46]$, and we set $\delta = 46 +17 = 63$. We can use Hoeffding's inequality\cite{hoeffding} to give confidence that $\langle \mo \rangle_{P^*} > 1$:
\be
\Pr(\langle \mo \rangle_{P^*} > 1) &= &1 - \exp(-2 M (\bar{\mo} - 1)^2 / \delta^2) \\
 &\approx& 1- 10^{-20}
\ee
For this problem, we were able to solve Eq.~\ref{sosprogram} in about thirty seconds on a contemporary laptop. Increasing $d$ does lead to tighter bounds, but, in this example, a relaxed bound suffices to rule out latent homophily with very high confidence.
Increasing $M$ also quickly increases confidence, making this technique useful for large datasets. 

This example shows that for a realistic influence model on a real social network, we are able to rule out latent homophily as the sole explanation for correlations. Although sociological studies identifying influence as a source of correlation do not typically test whether latent homophily could have explained the correlations\cite{obesity}, and despite the suggestion in \cite{cosma} that no such tests exist, this example demonstrates that we can realistically construct such tests and rule out latent homophily with very high confidence.

 \section{Related work}
 
We previously pointed out that a model like the one described in Eq.~\ref{algmodel} is a semi--algebraic set, that is, a set specified by a finite number of polynomial equalities and inequalities in the variables $\vc{x}\in \mathbb{R}^m,\vc{y}\in \mathbb{R}^n$, e.g. $f_1(\vc{x},\vc{y}) =0, \ldots, f_{l+1} (\vc{x},\vc{y}) \geq 0, \ldots, f_k(\vc{x},\vc{y}) \geq 0$. However, we do not directly observe the variables $\vc{x}$, so we are really solving the problem, given $\hat{\vc{y}} $, $\exists \vc{x}, f_1(\vc{x},\hat{\vc{y}})=0,\ldots$. This is considered the projection of our  semi--algebraic set onto the $\vc{y}$ variables only. The Tarski-Seidenberg theorem guarantees that the projection of a semi--algebraic set is, itself, a semi--algebraic set\cite{geigermeek}. That means we are guaranteed a representation of the set in terms of a finite number of polynomial equalities and inequalities involving $\vc{y}$ only. Converting from the first representation to the second is sometimes called \emph{quantifier elimination}. This is the approach used in \cite{geigermeek} to identify necessary and sufficient conditions for an observed distribution to be generated according to some latent variable graphical model. Unfortunately, the best known method for performing quantifier elimination, cylindrical algebraic decomposition, is doubly exponential in the number of variables, and is intractable even for simple models like Fig.~\ref{fig:chsh} or the instrumental inequality model\cite{kangtian}. 

\emph{Implicitization} is another algebraic technique that uses Gr\"obner bases to find the smallest \emph{algebraic variety} (set of polynomial equality statements) that contains a semi-algebraic set~\cite{geigermeek2,kangtian2}. Unfortunately, this approach has two limitations. First, it can only find equality constraints, which may not be sufficient for some models. For instance, in the CHSH experiment case, the smallest algebraic variety containing ``local hidden variable'' models will also contain quantum correlations; whereas, the CHSH inequalities rule out some quantum correlations\cite{thesis}. 
The second drawback of the implicitization approach is that its complexity depends on the size of the domains of all the model variables and, in principle, the size of the domain of the latent variable could be infinite.

A similar approach to implicitization also allows one to find equality constraints among observed variables in a latent variable graphical model and is given in \cite{tianpearl}. This approach only suffers one of implicitization's drawbacks, i.e. the domain of the latent variable could be infinite, but we are still restricted to equality constraints only, which may be insufficient.

In the Bayesian graphical model literature, the most studied example of hidden variable tests are the instrumental inequalities, which can be derived using linear programming techniques\cite{pearl} and were studied in greater detail in 
\cite{bonet}. 
In \cite{kangtian}, they considered the same general question as this paper, namely, a general method for identifying inequality constraints in models with hidden variables.  
Their approach leads to a specific necessary (but not sufficient) set of inequalities, but there is no way to produce tighter inequalities, or, as in our case, to produce an inequality optimized to rule out a given observation.

%\cite{sontag}

\section{Conclusion}

We have demonstrated an efficient method for constructing hidden variable tests that allow us to determine if an observed distribution is incompatible with a hidden variable model even if the domain of the hidden variables is infinitely large. 

In a machine learning or structure learning context, a typical approach to model selection compares a null model to an alternate model. One computes the likelihood of generating the data given the model and compares the ratio of likelihoods. This process may result in accepting a null hypothesis that could have been ruled out with high probability even before positing an alternate hypothesis. Conversely, one may find that an alternate model is more likely to explain some data, but there may be implications (legal, scientific, etc.) if you cannot rule out the possibility of explaining the data with the null hypothesis.

Studies in identifiability for Bayesian graphical models have either focused on independence relations, or, when considering non-independence constraints, they have concentrated on the particular case of the instrumental inequalities. 
Our method applies to a general class of models, and we have demonstrated its usefulness by applying it to the outstanding problem of ruling out latent homophily as the source of correlations in social networks.
Further work will explore results on some of the many commonly used models to which our method applies, such as HMMs.
Finding necessary \emph{and} sufficient conditions for arbitrary hidden variable models is another laudable goal, but the difficulty of doing so in the special case of Bell inequalities suggests it is an unrealistic one.

\subsection*{Acknowledgments}
We thank to Cosma Shalizi and Jennifer Neville for useful discussions while visiting ISI. This research was supported in part by the National Science Foundation
under grant No. 0916534 and US AFOSR MURI grant No. FA9550-10-1-0569.

\appendix
\section{Bell inequalities and graphical models}\label{a1}
In a quantum textbook describing the CHSH experiment, you will see the assumptions of local realism written in a form which is not obviously the same as the graphical model we show in Fig. 2. The connection between the two requires an argument like the one used in (Pearl, 2009, Sec. 8.2.2).  We will give a brief non-rigorous argument.

Suppose Alice has two measurement choices $X \in \{\pm1\}$, and the outcome is $A \in \{\pm 1\}$. In a quantum textbook, the assumption of local realism is that $A$ is a deterministic function of $X$ and some (possibly continuous) hidden variable $\lambda$, i.e. $A(X,\lambda) :  \{\pm1\} \times \mbox{dom}(\lambda) \rightarrow \{\pm1\}$. The same is the case for Bob's outcome, $B(Y,\lambda)$, with measurement choice $Y$. Then if we look at the probability of a certain outcome, 
\benn
Pr (A=a,B=b|X=x,Y=y) = \\
\int d\mu(\lambda) [A(X=x,\lambda)=a] [B(Y=y,\lambda)=b] ,
\eenn
where $\mu$ is some measure on $\lambda$.
Without loss of generality, we can decompose $\lambda$ into three parts, $\lambda_0, \lambda_A, \lambda_B$, so that $A(X,\lambda_0,\lambda_A)$ and $B(Y,\lambda_0,\lambda_B)$. That is, $A$ depends on some local hidden variable and some joint hidden variable that is shared with $B$. Now if we perform the integrals over $\lambda_A,\lambda_B$, and say that $\int d\mu(\lambda_A) [A(X=x,\lambda_0,\lambda_A)=a] \equiv P(A=a|X=x,R=\lambda_0)$, and similarly for $B$, we get  
 \benn
  Pr (A=a,B=b|X=x,Y=y) = \\
  \int d\mu(\lambda_0) P(A=a|X=x,R=\lambda_0) \times \\ P(A=a|X=x,R=\lambda_0),
  \eenn
 which has the form required, except with a continuous latent variable $\lambda_0$. Due to the finite state space (there are only 16 possible combinations of $a,b,x,y$) this can be written in terms of a discrete latent variable. 
 A similar argument holds in reverse; given $P(AB|XY)$ in terms of a graphical model in Fig. 2, one can construct a hidden variable and functions $A(X,\lambda),B(Y,\lambda)$, that reproduce the same statistics.
 
 \section{Strength of influence}\label{a2}
 What is the minimum amount of influence necessary to allow latent homophily to be ruled out? We provide an upper bound for the following special model.
 Consider a model of influence between Alice and Bob. Alice (Bob) may take a state $A^t (B^t) \in \pm1$ at each time step $t=1,\ldots,T$. Bob's state is chosen randomly so that $p(B^t= 1)=1/2$, and Alice is influenced with probability $\lambda$ to have the same value at $t+1$ as Bob did at time $t$, otherwise, Alice's state is random. 
We choose an observable,
\be\label{chooseO}
\mo = \frac1{T-1} \sum_{t=1}^{T-1} A^{t+1} B^t - \frac1{T-2} \sum_{t=1}^{T-2} B^t B^{t+2}.
\ee
For this model it is straightforward to derive $\langle \mo \rangle_{inf} = \lambda,$ regardless of $T$.
However, for an SLH model, it can be shown with standard algebraic optimization that for $T=4$, $\langle \mo \rangle_{SLH} \leq 0.376$, while for $T=6$, $\langle \mo \rangle_{SLH} \leq 0.302$. Therefore, if the probability of influence, $\lambda$, and therefore $\langle \mo \rangle$ exceeds these values, latent homophily can be ruled out as the sole explanation of the correlations. Interestingly, for a sequence of observations $T<4$, $\langle \mo \rangle_{SLH}$ is not bounded (except trivially by $1$). This is only an upper bound on the minimum influence need because, in principle, one could construct (e.g., using Eq.~\ref{sosprogram}) an observable, $\mo$ to rule out latent homophily for even smaller values of $\lambda$.
 
\bibliographystyle{plain}
{\bibliography{gversteeg}{}}

\begin{thebibliography}{10}

\bibitem{SM}
Code example is available at
  \url{http://drop.isi.edu/sites/default/files/users/gregv/soscode.pdf}.

\bibitem{anagnostopoulos}
Aris Anagnostopoulos, Ravi Kumar, and Mohammad Mahdian.
\newblock Influence and correlation in social networks.
\newblock In {\em In Proceedings of the 14th ACM SIGKDD}, pages 7--15. ACM,
  2008.

\bibitem{aral}
Sinan Aral, Lev Muchnik, and Arun Sundararajan.
\newblock Distinguishing influence-based contagion from homophily-driven
  diffusion in dynamic networks.
\newblock {\em PNAS}, 106(51):21544+, December 2009.

\bibitem{bonet}
Blai Bonet.
\newblock Instrumentality tests revisited.
\newblock In {\em In Proceedings of the 17th Conference on Uncertainty in
  Artificial Intelligence}, pages 48--55. Morgan Kaufmann, 2001.

\bibitem{obesity}
Nicholas~A. Christakis and James~H. Fowler.
\newblock The spread of obesity in a large social network over 32 years.
\newblock {\em The New England Journal of Medicine}, 357(4):370--379, July
  2007.

\bibitem{convergence}
Etienne De~Klerk and Monique Laurent.
\newblock Error bounds for some semidefinite programming approaches to
  polynomial minimization on the hypercube.
\newblock {\em SIAM Journal on Optimization}, 20(6):3104--3120, 2010.

\bibitem{geigermeek2}
D.~Geiger and C.~Meek.
\newblock Graphical models and exponential families.
\newblock {\em Proceedings of the 14th Conference on Uncertainty in Artificial
  Intelligence}, 1998.

\bibitem{geigermeek}
D.~Geiger and C.~Meek.
\newblock Quantifier elimination for statistical problems.
\newblock {\em Proceedings of the 15th Conference on Uncertainty in Artificial
  Intelligence}, 1999.

\bibitem{hoeffding}
W.~Hoeffding.
\newblock Probability inequalities for sums of bounded random variables.
\newblock {\em Journal of the American Statistical Association},
  58(301):13--30, 1963.

\bibitem{kangtian}
Changsung Kang.
\newblock Inequality constraints in causal models with hidden variables.
\newblock In {\em In Proceedings of the 17th Annual Conference on Uncertainty
  in Artificial Intelligence}, pages 233--240, 2006.

\bibitem{kangtian2}
Changsung Kang and Jin Tian.
\newblock Polynomial constraints in causal bayesian networks.
\newblock In {\em In Proceedings of the 17th Annual Conference on Uncertainty
  in Artificial Intelligence}, 2007.

\bibitem{neville}
Timothy La~Fond and Jennifer Neville.
\newblock Randomization tests for distinguishing social influence and homophily
  effects.
\newblock In {\em WWW '10: Proceedings of the 19th international conference on
  World Wide Web}, pages 601--610. ACM, 2010.

\bibitem{lermandigg}
Kristina Lerman and Rumi Ghosh.
\newblock Information contagion: an empirical study of spread of news on digg
  and twitter social networks.
\newblock In {\em Proceedings of 4th International Conference on Weblogs and
  Social Media (ICWSM)}, May 2010.
\newblock Data available, http://www.isi.edu/\~{}lerman/.

\bibitem{nielsen}
{Nielsen, Michael A.} and {Chuang, Isaac L.}
\newblock {\em {Quantum Computation and Quantum Information}}.
\newblock {Cambridge University Press}, {October} 2000.

\bibitem{parrilo}
Pablo~A. Parrilo.
\newblock Semidefinite programming relaxations for semialgebraic problems.
\newblock {\em Math. Program.}, 96(2, Ser. B):293--320, 2003.

\bibitem{pearl}
Judea Pearl.
\newblock {\em Causality: Models, Reasoning and Inference}.
\newblock Cambridge University Press, New York, NY, USA, 2009.

\bibitem{peres}
A.~Peres.
\newblock {\em Quantum Theory: Concepts and Methods}.
\newblock Kluwer Academic Publishers, 1993.

\bibitem{pitowsky}
Itamar Pitowsky.
\newblock Correlation polytopes: their geometry and complexity.
\newblock {\em Math. Program.}, 50(3):395--414, 1991.

\bibitem{sostools}
Stephen Prajna, Antonis Papachristodoulou, Peter Seiler, and Pablo~A. Parrilo.
\newblock Sostools and its control applications.
\newblock In Didier Henrion and Andrea Garulli, editors, {\em Positive
  Polynomials in Control}, pages 273--292. Springer Berlin / Heidelberg, 2005.

\bibitem{convexanalysis}
Ralph~T. Rockafellar.
\newblock {\em Convex Analysis}.
\newblock {Princeton University Press}, 1996.

\bibitem{cosma}
Cosma~R. Shalizi and Andrew~C. Thomas.
\newblock Homophily and contagion are generically confounded in observational
  social network studies.
\newblock {\em arxiv:1004.4704}, 2010.

\bibitem{thesis}
Greg~Ver Steeg.
\newblock {\em Foundational aspects of nonlocality}.
\newblock PhD thesis, California Institute of Technology, 2009.

\bibitem{sedumi}
{Sturm, J. F.}
\newblock Using sedumi 1.02, a {MATLAB} toolbox for optimization over symmetric
  cones.
\newblock {\em {Optimization Methods and Software}}, 11--12:625+, 1999.

\bibitem{tianpearl}
J~Tian and J~Pearl.
\newblock On the testable implications of causal models with hidden variables.
\newblock In {\em In Proceedings of the Conference on Uncertainty in Artificial
  Intelligence}, pages 519--527. Morgan Kaufmann Publishers, 2002.

\end{thebibliography}

\end{document}